# Multi-view deep learning based molecule design and structural optimization accelerates the SARS-CoV-2 inhibitor discovery

Chao Pang[1,2], Yu Wang[1,2], Yi Jiang[1,2], Ruheng Wang[1,2], Ran Su[3*], and Leyi Wei[1,2*]

[1]School of Software, Shandong University, Jinan, China;

[2]Joint SDU-NTU Centre for Artificial Intelligence Research (C-FAIR), Shandong University, Jinan, China;

[3]College of Intelligence and Computing, Tianjin University, Tianjin, China;

[*]Corresponding authors:

R.S.: ran.su@tju.edu.cn

L.W: weileyi@sdu.edu.cn

## Abstract

In this work, we propose MEDICO, a **M**ulti-vi**E**w **D**eep generative model for molecule generation, structural optimization, and the SARS-CoV-2 **I**nhibitor dis**CO**very. To the best of our knowledge, MEDICO is the first-of-this-kind graph generative model that can generate molecular graphs similar to the structure of targeted molecules, with a multi-view representation learning framework to sufficiently and adaptively learn comprehensive structural semantics from targeted molecular topology and geometry. We show that our MEDICO significantly outperforms the state-of-the-art methods in generating valid, unique, and novel molecules under benchmarking comparisons. In particular, we showcase the multi-view deep learning model enables us to generate not only the molecules structurally similar to the targeted molecules but also the molecules with desired chemical properties, demonstrating the strong capability of our model in exploring the chemical space deeply. Moreover, case study results on targeted molecule generation for the SARS-CoV-2 main protease ($M^{pro}$) show that by integrating molecule docking into our model as chemical priori, we successfully generate new small molecules with desired drug-like properties for the $M^{pro}$, potentially accelerating the *de novo* design of Covid-19 drugs. Further, we apply MEDICO to the structural optimization of three well-known $M^{pro}$ inhibitors (N3, 11a, and GC376) and achieve ~88% improvement in their binding affinity to $M^{pro}$, demonstrating the application value of our model for the development of therapeutics for SARS-CoV-2 infection.

# Introduction

*De novo* drug design is a time-consuming and expensive research project. Developing new drugs costs an average of 2.6 billion U.S. dollars and takes researchers 12 years, and this number is still growing[1,2]. In drug discovery, the rapid and accurate discovery of lead compounds is a very important topic. The molecular space of drug-like molecules is very huge (estimated to be in the order of $10^{60}$)[3], generating drug-like molecules still remains challenging. With the improvement in calculation ability, attempts have been made to use computer to assist in drug discovery, for instance, high-throughput screening[4], pharmacophore modeling[5] and virtual screening[6]. Although these methods can accelerate drug discovery, they still have some drawbacks, like limited searchable chemical space and time-consuming calculation process. In recent years, deep learning has achieved great success in many fields, such as face recognition[7], image generation[8], and natural language processing (NLP)[9]. Deep generative models have been extensively introduced into the *de novo* molecule design, and successfully accelerated drug discovery to some extent. They are data-driven methods that enable molecule generation via capturing the underlying rules of data distribution and sampling data from this distribution. According to the information used for modeling in the deep generative approaches, they are generally divided into the three categories: (1) Sequence-based approaches, (2) Topology-based approaches, and (3) Geometry-based approaches.

Sequence-based approaches use text-like data such as Simplified Molecular Input Line Entry System (SMILES)[10] as molecule representations, and then map molecules into the latent space to generate new molecules[11]. For instance, Ertl et al. presented a method to generate molecules using a long short-term memory (LSTM) neural network[12]. Grisoni et al. developed BIMODAL, a bidirectional generative RNNs for bidirectional molecule design by alternating learning[13]. Dai et al. proposed a deep Variational Auto-Encoder (VAE) model namely SDVAE, which integrates semantic restriction in syntax tree generator, allowing the model effectively generate both syntactical and semantically valid molecules[14]. However, these sequence-based approaches suffer from some intrinsic limitations. For example, using SMILES as molecule representations cannot capture the molecular structural similarity; on the other hand, it easily losses the topological information of molecules as well, thus greatly impacting the model performance.

To solve above problems, topology-based approaches are proposed. They utilize molecule graphs as molecule representations, train the deep generative models, and enable directly generate the molecule graphs[15]. Li et al. proposed MolMP, a new de novo molecular design framework based on a type of sequential graph generator. However, the sequential graph generation (i.e. node-by-node) suffers from random permutation of nodes, which affects validity of generated molecules. Consequently, Jin et al. developed a Junction Tree Variational Autoencoder (JTVAE) model by assembling substructures of molecules to improve the node-by-node generation, generating 100% valid molecules[16]. Madhawa et al. proposed GraphNVP, a flow-based molecular graph generation model through graph structure and node label[17]. However, the methods cannot encode 3D geometry of molecules in molecule design. It has been proven in studies that new active molecules and binding sites can be successfully predicted using molecular geometry, which cannot be done by sequence or topology[18]. This demonstrates that molecular geometry plays an important role in biochemical function and activity prediction.

To utilize geometric information of molecules, researchers proposed a series of geometry-based approaches. Skalic el at. proposed LigDream, a voxel-based model using 3D molecule shape and molecule pharmacophoric features to generate molecules[19]. Imrie et al. developed a novel deep generative model called DeLinker that takes two fragments or partial structures and designs a molecule utilizing the relative distance and orientation to link them[20]. Li et al. proposed GEOM-CVAE a constrained variational autoencoder based on image-like geometric representation for molecular generation[21]. However, all of above methods have problems in utilizing geometric information. The 3D shape used by LigDream is generated from an autoencoder, which in fact does not introduce additional useful information[19]. DeLinker requires additional decomposition of molecular structures to obtain fragments, which remarkably increases the computational complexity[20]. The geometric representation used in GEOM-CVAE does not satisfy rotation and translation invariance, leading to the lack of specificity in molecule representations[21]. To this end, how efficiently use geometric information is crucial for molecule generation.

Unfortunately, all three types of molecular generative methods mentioned above have the following inevitable drawbacks. (1) Sequence and topology data do not contain structural information of the molecule, while the relationship between atoms in molecule cannot be clearly represented using only geometry. Liu et al. demonstrate that the joint representation of topology and geometry is superior to an ordinary molecular graph representation through experimental and theoretical analysis[22]. (2) Previous studies demonstrate that structurally similar molecules are likely similar in terms of their potential as drugs and other properties[23], which is crucial in drug design and discovery. However, the existing methods ignore the importance of molecular structures in model construction, leading to their poor ability of generating new molecules structurally similar to targeted molecules. (3) Compared with the compounds generated by existing unconstrained models, the ones generated by protein-constrained models have more clear advantages in maintaining the desirable binding mode as well as favorable docking scores[24].

In this work, we propose MEDICO, an innovative deep generative model for *de novo* molecule design. Our main contributions can be summarized as follows: (1) We propose a multi-view representation learning framework to explore discriminative information from both the molecule topology and geometry. With the learnt joint representations, MEDICO can generate molecules that are structurally similar to the given seed molecule, which is crucial in the optimization problem of lead compounds. To the best of our knowledge, MEDICO is the first-of-this-kind graph generative model that can generate molecular graphs similar to the structure of a given molecule. (2) We propose a flow-based deep generative model for topology processing and generation, and a message passing neural network (MPNN) for geometry processing. Flow-based models can memorize and accurately reconstruct the input information and generate novel, unique and effective molecules. MPNN is responsible for processing molecular geometry into physically meaningful molecule representations. These allow our model to explore the chemical space more deeply. (3) We introduce molecular docking results of the main protease ($M^{pro}$, also named 3CL protease) of severe acute respiratory syndrome coronavirus 2 (SARS-CoV-2) as chemical priori, which is essential for generating molecules with potential as Covid-19 drugs[25]. (4) To explore the effectiveness of our model in practical applications, we apply MEDICO to property optimization and molecular optimization tasks. Experimental results demonstrate that MEDICO is able to generate molecules with desirable properties and binding mode, illustrating that MEDICO has great application value in downstream tasks and potential for extension to other tasks.

## Methods and materials

**Overview of MEDICO**

The overall framework of our MEDICO is shown in **Figure 1**. As can be seen in **Figure 1a**, MEDICO consists of three modules: (1) Data pre-processing for SARS-CoV-2, (2) Multi-view representation learning, and (3) Flow-based molecule generation. In data pre-processing for SARS-CoV-2 module, we first calculate docking scores for all molecules in the dataset with $M^{pro}$ of SARS-CoV-2. These docking scores are used as chemical priori information, giving each molecule a weight. The raw data of molecule SMILES and XYZ files are preprocessed into 2D topology and 3D geometry of molecules, respectively. In multi-view representation learning module, to process topology and geometry of molecules, we use flow-based model and MPNN, respectively. The Flow-based model encodes 2D topology as latent vectors in high-dimensional space (see **Figure 1b**) while MPNN encodes 3D geometry into physically meaningful representations based on spherical coordinate system (SCS) (see **Figure 1c**). Afterward, MEDICO combines topological representations with geometric representations as a joint representation of the molecule. Finally, in flow-based molecule generation module, flow model's reverse process transforms the joint representation into a molecular graph. The details of our MEDICO regarding the three main modules are described below.

**Data pre-processing for SARS-CoV-2**

*Molecular docking for SARS-CoV-2.* Molecular docking is a key tool for computer-aided drug design. A successful docking method can efficiently search high-dimensional spaces and use a scoring function to correctly rank candidate molecules[26]. General molecular docking procedures are time-consuming, and most of them require manual determination of docking positions and docking pockets[27], so they cannot be used in batch processes for docking a large number of ligands. To meet our need for fast and automatic screening of molecules in the dataset, we use AutoDock Vina 1.2.0, one of arguably the fastest and most widely used open-source programs for molecular docking[28]. By using AutoDock Vina 1.2.0, we can achieve fast and valid molecular docking.

We choose $M^{pro}$ of SARS-CoV-2 as the protein receptor in our molecular docking. It is one of the most important antiviral drug targets of Covid-19[29]. We use the crystal structure of SARS-CoV-2 $M^{pro}$ in complex with N3 to a resolution of 2.1 Å (PDB ID: 6LU7)[25]. To perform molecular docking, we first remove the N3 molecule from the crystal structure and then use its location as a pocket for molecular docking.

In terms of ligands for molecular docking, small molecule files require pre-processing before they can be used as the final ligands for molecular docking. We convert small molecules into PDB files with Open Babel, which is a powerful tool for converting chemical data formats[30]. In addition, we use PDB2PQR[31] to calculate the electrostatic potential of the protein surface and convert PDB files into PQR files.

Finally, we apply the negative value of the binding energy obtained by molecular docking as a basis to assign a weight $w$ for each molecule in the training set in order to assess the importance of different molecules. Let $\alpha$ ($\alpha > 0$) denote the negative value of the binding energy, then the probability of molecule $i$ being selected during each epoch of training is:

$$P_{pick}^i = w = \frac{\alpha_i - \alpha_{min}}{\alpha_{max} - \alpha_{min}}, \#(1)$$

where $\alpha_i$ is the negative value of binding energy of molecule $i$; $\alpha_{max}$ and $\alpha_{min}$ represent the maximum and minimum values of the calculated binding energy, respectively.

***Representation of 2D topology.*** For SMILES, we first transform it into a molecular graph, and then divide the molecule into atom tensor and chemical bond tensor[32]. More specifically, a molecule $\mathcal{M}$ can be represented as a collection of an atomic tensor $\mathcal{A} \in \{0, 1\}^{n \times l}$ and a bond tensor $\mathcal{B} \in \{0, 1\}^{n \times n \times m}$, where $n, l, m$ denote the number of atoms, atom type, and chemical bond type, respectively. $\mathcal{A}[i, p] = 1$ represents the $i$-th atom with atom type $p$ and $\mathcal{B}[i, j, q] = 1$ represents the type of chemical bond between the $i$-th and $j$-th atoms is $q$. Subsequently, the atom tensor and bond tensor are used as inputs for Graph conditional flow and Glow in flow model.

***Representation of 3D geometry.*** The molecular 3D geometry in the dataset is usually represented as coordinate points of each atom in 3D Cartesian coordinates. However, the representation cannot satisfy the rotation and translation invariance. For example, if we rotate a molecule by 60 degrees around the z-axis, or translate it by one-unit length in the positive direction of the y-axis, we will get a representation that looks completely different from the original molecule while they are the same. The above problem can be solved by using SCS to represent molecules. In SCS, the position of an atom is determined by a triad $(r, \theta, \varphi)$, where $r, \theta, \varphi$ denote the radial distance, polar angle, and the azimuthal angle, respectively. Although the representation derived directly from SCS can represent a molecular conformation completely, it requires edge-based 2-hop information of computational complexity $O(nk^3)$, where $n$ is the number of atoms, and $k$ denotes the average number of neighboring atoms[33]. Therefore, similar to Liu et al.[33], we use spherical message passing (SMP). In detail, the 3D geometry of a molecule $\mathcal{G}$ with $n$ atoms and $m$ edges can be represented as a 4-tuple $\mathcal{G} = (\boldsymbol{u}, \mathcal{V}, \mathcal{E}, \mathcal{W})$, where $\boldsymbol{u} \in \mathbb{R}^{d_u}$ denotes the global feature vector of $\mathcal{G}$, $\mathcal{V} = \{\boldsymbol{v}_i\}, i \in [1, n]$ denotes the set of atom feature vectors, $\mathcal{E} = \{\boldsymbol{e}_j, r_j, s_j\}, j \in [1, m]$ denotes the set of edge feature vectors, $r_j, s_j$ represent the index of the message receiving atom and the message sending atom respectively. $\mathcal{W} = \{(x, y, z)_i\}, i \in [1, n]$ denotes the set of 3D Cartesian coordinate points of atoms.

**Multi-view molecule representation learning**

***The flow model framework in processing of 2D topology.*** A flow model is a generative model consisting of a series of reversible transformations stacked $f_\Theta = f_N \circ \ldots \circ f_2 \circ f_1$. It can map the real data point $x \in \mathbb{R}^D$ to the hidden variable $z \in \mathbb{R}^D$, where $x \sim P_X(x)$, $z \sim P_Z(z)$, $P_X$ is the complex distribution that the real data $x$ obeys, $P_Z$ can be any distribution, generally, we define it as a simple distribution, such as a Gaussian distribution. $P_X$ can be obtained by change of variable formula as follows:

$$P_X(x) = P_Z(z) \left| \det\left(\frac{\partial z}{\partial x}\right) \right| \#(2)$$

where $z = f_\Theta(x)$. We sample a random point $z$ from the simple distribution $P_Z$, and use $x = f_\Theta^{-1}(z)$, then we get the newly generated real data point $x$. Let $L_n = f_n(L_{n-1})$ be the hidden variable of $x$ after $n$ transformations, $n = 1, 2, ..., N$, $L_0 = x$, the log-likelihood of Equation (1) can be written as:

$$\log P_X(x) = \log P_Z(z) + \log \left| \det\left(\frac{\partial z}{\partial x}\right) \right|$$
$$= \log P_Z(z) + \sum_{n=1}^{N} \log \left| \det\left(\frac{\partial f_n}{\partial L_{n-1}}\right) \right| \#(3)\#\#$$

However, since the Jacobi determinant of transformation $f$ is intractable, we design the transformation $f$ to reduce the cost of computing the Jacobi determinant. Dinh et al. proposed the affine coupling layer[34,35], which solves the problem of computational complexity of Jacobi determinant. Both $z$ and $x$ are split into two parts: the front $d$ dimension and the $d+1$ to $D$ dimension. The process from $z$ to $x$ is to copy the first $d$ dimensions of $z$ directly to the first $d$ dimensions of $x$. The $d+1$ to $D$ dimensions of $z$ is calculated using a flexible and tractable bijective function by stacking a sequence of simple bijections[35]:

$$z_{1:d} = x_{1:d}$$
$$z_{d+1:D} = x_{d+1:D} \odot \exp(s(x_{1:d})) + t(x_{1:d}) \#(4)$$

where $s$ and $t$ stand for scale and translation functions, $\odot$ is the Hadamard product. The inverse of the above equation is:

$$x_{1:d} = z_{1:d}$$
$$x_{d+1:D} = ((z_{d+1:D}) - t(z_{1:d})) \odot \exp(-s(z_{1:d})) \#(5)$$

The Jacobian of the affine coupling layer is:

$$J_f(z) = \begin{bmatrix} \mathbb{I}_d & 0 \\ \frac{\partial z_{d+1:D}}{\partial x_{1:d}^T} & \text{diag}(\exp[s(x_{1:d})]) \end{bmatrix} \#(6)$$

Thus, the Jacobian determinant can be computed efficiently by $J_f(z) = \text{diag}(\exp[s(x_{1:d})])$.
In the training phase, since we divide the molecule into two parts, atom tensor $\mathcal{A}$ and bond tensor $\mathcal{B}$, $P(\mathcal{M})$ can be divided into two parts too:

$$P(\mathcal{M}) = P((\mathcal{A}, \mathcal{B})) \approx P(\mathcal{A}|\mathcal{B}; \theta_{\mathcal{A}|\mathcal{B}}) P(\mathcal{B}; \theta_\mathcal{B}) \#(7)$$

where $\theta_{\mathcal{A}|\mathcal{B}}$ is the parameter learned by the graph conditional flow model conditioned on the bond tensor $\mathcal{B}$. $\theta_\mathcal{B}$ is the parameter learned by the Glow model. These two parameters can be obtained by maximizing the following equation:

$$\underset{\theta_{\mathcal{A}|\mathcal{B}}, \theta_\mathcal{B}}{\text{argmax}} \mathbb{E}_{(\mathcal{A},\mathcal{B}) \sim P(\mathcal{M})} [\log P(\mathcal{A}|\mathcal{B}; \theta_{\mathcal{A}|\mathcal{B}}) + \log P(\mathcal{B}; \theta_\mathcal{B})] \#(8)$$

The workflow of the flow model is detailed in **Figure 1b**.

***The MPNN framework in processing of 3D geometry.*** After getting 3D geometry representation

$\mathcal{G} = (\boldsymbol{u}, \mathcal{V}, \mathcal{E}, \mathcal{W})$ of a molecule, the feature vectors $\boldsymbol{u}, \boldsymbol{v} \in \mathcal{V}, \boldsymbol{e} \in \mathcal{E}$ of this 4-tuple can be updated by a message passing network:

$$\boldsymbol{e}'_j = g^e\left(\boldsymbol{e}_j, \boldsymbol{v}_{r_j}, \boldsymbol{v}_{s_j}, \mathcal{E}_{s_j}, h^{w,e}(\{(x,y,z)_\gamma\})\right), \gamma = r_j \cup s_j \cup \Gamma_{s_j}$$

$$\boldsymbol{v}'_i = g^v\left(\boldsymbol{v}_i, h^{e,v}(\mathcal{E}'_i)\right)$$

$$\boldsymbol{u}' = g^u(\boldsymbol{u}, h^{v,u}(\mathcal{V}')) \#(9)$$

where $g$ is a message update function, $h$ is a message aggregate function and $\Gamma_{s_j}$ is the set of atoms connected to the edges in $\mathcal{E}_{s_j}$. The coordinate system conversion and message process of SphereNet are shown in **Figure 1c** and the process of message update and aggregation of SphereNet are shown in **Supplementary Figure S1**. In terms of network architecture, SphereNet consists of an input block, multiple interaction blocks, and an output block. The input block is responsible for constructing the initial messages for the edges, the interaction block is responsible for updating the messages, and the output block aggregates all inputting messages together.

Since $(r, \theta, \varphi)$ is not a meaningful physical representation, they cannot be directly involved in the message passing process. Therefore, we use some basis functions which include spherical Bessel basis and spherical harmonics[36,37] to convert $(r, \theta, \varphi)$ into physical-based representations $\Psi(r), \Psi(r, \theta)$ and $\Psi(r, \theta, \varphi)$. The input block takes only the distance representation $\Psi(r)$ as the input. The interaction block's inputs include all three representations. After several message updating processes, the final obtained global feature vector $\boldsymbol{u}^*$ will be used as the representation of 3D geometry.

*Acquisition of joint representation.* By inputting the molecular graph into the trained flow model, we can obtain the latent vector that obeys the Gaussian distribution $\boldsymbol{z}_{\mathcal{A}|\mathcal{B}} \in \mathbb{R}^{d_{\mathcal{A}|\mathcal{B}}}, \boldsymbol{z}_{\mathcal{B}} \in \mathbb{R}^{d_{\mathcal{B}}}, d_{\mathcal{A}|\mathcal{B}} = n \times l, d_{\mathcal{B}} = n \times n \times m$. Let $\boldsymbol{z}_{\mathcal{M}} = \boldsymbol{z}_{\mathcal{A}|\mathcal{B}} \| \boldsymbol{z}_{\mathcal{B}} \in \mathbb{R}^{d_{\mathcal{A}|\mathcal{B}} + d_{\mathcal{B}}}$, where $\|$ stands for concate operation. Molecular conformational data can be encoded by SphereNet[33] to obtain feature vectors of molecular 3D geometry $\boldsymbol{u}^* \in \mathbb{R}^{d_{\mathcal{A}|\mathcal{B}} + d_{\mathcal{B}}}$. We define the fusion loss as follows:

$$\mathcal{L}_{\text{fus}} = \mathbb{E}_{\mathcal{M} \sim P_{\mathcal{M}}} \|\boldsymbol{z}_{\mathcal{M}} - \boldsymbol{u}^*\|_2 \#(10)$$

$\boldsymbol{z}_{\mathcal{M}}$ contains the 2D topological information of the molecule, while $\boldsymbol{u}^*$ contains the 3D geometric information of the molecule. We train SphereNet with $\mathcal{L}_{\text{fus}}$ loss function to produce a joint representation $\boldsymbol{u}_{\mathcal{M}}$ which is similar to $\boldsymbol{z}_{\mathcal{M}}$. Besides, we can control the proportion of 2D topology and 3D geometry contained in the final molecular representation $\boldsymbol{u}_{\mathcal{M}}$ by setting the number of training epochs.

**Flow-based molecule generation**

After obtaining the joint representation $\boldsymbol{u}_{\mathcal{M}}$ of 2D topology and 3D geometry, we consider it as a latent vector of molecular graph with 3D geometry $\boldsymbol{z}'_{\mathcal{M}} \in \mathbb{R}^{d_{\mathcal{A}|\mathcal{B}} + d_{\mathcal{B}}}$. Next, we split it into $\boldsymbol{z}'_{\mathcal{A}|\mathcal{B}} \in \mathbb{R}^{d_{\mathcal{A}|\mathcal{B}}}$ and $\boldsymbol{z}'_{\mathcal{B}} \in \mathbb{R}^{d_{\mathcal{B}}}$, then use them as input data for reverse graph conditional flow and reverse Glow. These two models are the inverse processes of the flow model trained during 2D topology processing.

From equation (4), we have:

$$\mathcal{A}_{1:d} = z_{\mathcal{A}_{1:d}|\mathcal{B}}$$
$$\mathcal{A}_{d+1:D} = \left(\left(z_{\mathcal{A}_{d+1:D}|\mathcal{B}}\right) - t_\beta\left(z_{\mathcal{A}_{1:d}|\mathcal{B}}\right)\right) \odot exp\left(-s_\beta\left(z_{\mathcal{A}_{1:d}|\mathcal{B}}\right)\right) \#(11)$$

$$\mathcal{B}_{1:d} = z_{\mathcal{B}_{1:d}}$$
$$\mathcal{B}_{d+1:D} = \left(\left(z_{\mathcal{B}_{d+1:D}}\right) - t_\beta\left(z_{\mathcal{B}_{1:d}}\right)\right) \odot exp\left(-s_\beta\left(z_{\mathcal{B}_{1:d}}\right)\right) \#(12)$$

where $\beta$ is our model trained parameter. In addition, we use the Sigmoid function in practice instead of the exponential function.

Because valid molecules must conform to the rules of chemical valency, we add a chemical valency check module to the generation process. With this module turned on, molecules that do not match chemical valence rules will be filtered out.

**Datasets**

In this study, we use a public benchmark dataset namely QM9[38] for our experiments. QM9 consists of 134k stable small organic molecules of CHONF and their computed geometrical, energetic, electronic, and thermodynamic properties. These molecules correspond to a subset of all 133,885 species with up to nine heavy atoms (CONF) in the 166 billion organic molecules of the GDB-17 chemical universe[38,39].

**Baseline methods**

We choose the following methods as the comparison baseline: (1) VAE-based methods: CGVAE[40] and JTVAE[16] (2) Reinforcement learning (RL)-based method: GCPN[41] (3) Flow-based methods: GraphNVP[17] and GraphAF[42]. All these baseline methods are run with their official codes and default parameters.

**Performance evaluation metrics**

To evaluate the performance of our model in terms of molecule generation, we use three categories of evaluation metrics.

**(1)** *Metrics of the set of generated molecules*. **Validity** is the percentage of generated molecules that do not violate the rules of chemical valency or have syntactic problems. **Novelty** is the percentage of generated molecules that have not appeared from the training dataset. **Uniqueness** is the percentage of unique molecules in generated molecules.

**(2)** *Metrics to evaluate the similarity of the generated molecule to the original molecules*. **Tanimoto similarity**[43] measures the similarity calculated between topological fingerprints using Tanimoto coefficient. **Fraggle similarity** is used to evaluate the structural similarity of intermolecular fragments. **MACCS similarity** calculates the similarity between MACCS (Molecular Access System)[44] and a substructure keys-based fingerprint with 166 structural keys.

**(3)** *Metrics for evaluating the inhibitory effect of molecules on SARS-CoV-2*. We use RDKit[45] to generate the conformation corresponding to SMILES and then dock it to M$^{pro}$ of SARS-CoV-2, and the negative value of the binding energy obtained is used as the **docking score**, which can be used to measure the binding affinity between the ligand and the receptor protein.

# Results and discussions

**MEDICO outperforms the state-of-the-art methods in terms of valid, novel and unique molecule generation**

For the random generation task, we use the widely used evaluation metrics **Validity**, **Novelty** and **Uniqueness** to evaluate the performance of our model and baselines. In addition, some models incorporate a valence check module to ensure that all molecules generated are chemically valid. To compare the real performance of the models, we follow GraphAF and compare the percentage of valid molecules generated by the models without the chemical valence check, denoted as **Validity w/o check** in **Table 1**. Note that in the random generation task, MEDICO only uses 2D topology of molecules while 3D geometry of molecules and MPNN model are not used. For the baseline models, we follow their default parameters and settings for training and generation. We exploit each model to generate 10,000 molecules for performance comparison.

**Table 1** shows the performance of our model MEDICO vs. baselines on QM9. The experimental results show that MEDICO achieves competitive performances on all four metrics. MEDICO can achieve 100% validity rate in the case of using valency check and 94.06% validity rate without the use of valency check. For comparison, GCPN can only achieve 20% without valence check, while other two flow-based models GraphNVP and GraphAF also have a lower validity rate than our MEDICO, achieving 50.86% and 46.30%, respectively. In terms of novelty and uniqueness, MEDICO also performs well, reaching 95.08% and 99.45%. The results demonstrate that our MEDICO could generate valid, novel and unique molecules for random molecule generation. **Supplementary Figure S2** shows some examples of molecules randomly generated by MEDICO.

**MEDICO can generate new molecules structurally similar to seed molecules**

Molecules with similar structures are more likely to have similar physicochemical properties, and the fact that the generated molecules obey a similar distribution to those in the dataset further ensures that the generated molecules are more likely to be similar in structure and properties to those in the dataset. Therefore, the above two points are crucial for molecular generation models. Compared to other generation models, MEDICO can generate molecules with similar structures to the original molecule. This may be because our model uses 3D geometry of the molecule, while others do not. To explore the importance of 3D geometry of the molecule in molecule generation, we first randomly choose molecules from the test set as seed molecules, generate the molecules with and without their 3D geometry respectively, and then calculate the similarity between generated molecules and seed molecules using Fraggle similarity. To ensure the generative power of our model, we add 20% random noise to the joint representation generated by SphereNet (In this way our generated molecules can be more diverse. In all the molecule generation processes below, the same 20% random noise is added if not stated). **Figure 2** illustrates some seed molecules and corresponding molecules generated based on them. The value of Fraggle similarity between the generated molecule and its seed molecule is marked below each generated molecule. In the case of MEDICO using 3D geometry for generation, it can be seen that the molecules generated by MEDICO are very similar to seed molecules in both general frame and local structure. The calculated Fraggle similarity scores also demonstrate the high similarity of the MEDICO-generated molecules to seed molecules. At the same time, the generated molecules retain distinguishing substructures of seed molecules, such as ternary, pentacyclic and heptacyclic rings. However, when only using 2D topology, the similarity between the generated molecule and the seed molecule is significantly reduced. Moreover, many of the molecules generated have some unreasonable structures, such as multiple small rings combined together. This shows that 3D geometry is crucial in terms of structure-oriented molecule generation.

On the other hand, the molecules generated by MEDICO have a similar distribution to those in the original dataset. We first calculate the Morgan fingerprint[46] of the molecule and then use the t-SNE algorithm[47] to reduce the dimensionality of the Morgan fingerprint and visualize it in **Figure 3a**. In the t-SNE diagram, every dot denotes a molecule and two distance-close dots tend to have similar structures. Specifically, we select two pairs of points in the figure and show their structures. Furthermore, we choose four properties of molecules to observe their distributions of our generated molecules and molecules in the dataset. These four properties are Molecular weight, plogP[48], QED[49] and SA (Synthetic accessibility)[50]. Molecular weight can reflect the size of a molecule to a certain extent. The property plogP can reflect the water solubility of molecules. QED is for quantifying drug-like properties of molecules, and the value of SA shows whether the molecule is chemically easy to synthesize. The obtained violin diagrams are shown in **Figure 3b**. From **Figure 3a**, it is clear that the molecules in the dataset are well mixed with the generated molecules in the two-dimensional plane. This indicates that the distribution of molecular fingerprints of molecules generated by MEDICO is very similar to those in the dataset. From the violin diagram, it can be seen that for the four commonly used molecular properties, the distribution of the properties of generated molecules is close to that of the molecules in the dataset, indicating that the generated molecules also have similar distributions of molecular properties to those in the dataset.

**Optimization for docking score of $M^{pro}$**

To verify whether the model can generate molecules with higher docking scores, we randomly select 1000 molecules from the test set, use their 3D geometry to generate 1000 new molecules, and calculate their docking scores with $M^{pro}$ to explore the binding potential by using molecular docking on the generated molecules. For a fair comparison, instead of using the molecular conformation in QM9, we use the conformation generated by RDKit. The calculation of molecular docking still uses AutoDock Vina 1.2.0. The experimental results show that the average docking score of the seed molecules in the dataset is 4.217, while that of the generated molecules is 4.248. This indicates that MEDICO can indeed generate new molecules with a high docking score of $M^{pro}$.

Among these generated molecules, some of them achieve a high docking score boost. We select two examples of them and visualize them using Discovery Studio software[51]. The 2D and 3D schematics of docking results are shown in **Figure 4**. It can be seen that the generated molecule retains some substructures of the original molecule while other parts are changed to bind to more amino acid residues of the target protein. Thus, their docking score with $M^{pro}$ is substantially increased.

**The significant impact of 3D geometry on molecular generation**

To further explore the role of 3D geometry in MEDICO, we conduct three ablation experiments in each of the tasks mentioned above. In addition, in the molecular docking experiments, we also explore the effects of 3D geometry and molecular docking module on the experimental results respectively.

*Random molecule generation*. In the random generation task, we add different proportions of noise to the joint representation of 2D topology and 3D geometry to explore the effect of 3D geometry on random generation. From no noise added to all noise, we set up a set of experiments at 0.2 scale intervals, and the experimental results are shown in **Figure 3c** and **Supplementary Table S2**. The results show that the novelty and uniqueness metrics are basically unchanged in the six groups of experiments, indicating that 3D geometry does not significantly affect the novelty and uniqueness of the generated molecules. And the validity with no check increases significantly with the increase of the scale of random noise, which indicates that the additional introduced 3D geometry will impose a constraint on the model generation process and limit the molecular search space. Another possible part of the reason is the inevitable error in translating the 3D geometry of the molecule into latent vectors encoding 2D topology. Although introducing 3D geometry might limit the search space, it is necessary for our MEDICO to generate new molecules with similar structures to the given seed molecule.

*Structurally similar molecule generation.* To explore the effect of 3D geometry on the generation of similar molecules, we randomly select 1,000 molecules from the dataset, then use their 3D geometry (without noise) and random noise to generate 1,000 new molecules respectively. Finally, we evaluate the similarity of the generated molecules to the seed molecules using three similarity metrics, Tanimoto, Fraggle and MACCS. As a comparison, we randomly select 2,000 molecules in the dataset and calculated the similarity between the first half of the molecules and the second half. **Table 2** shows the mean values of the scores of three similarity metrics for these three generation conditions, while **Figure 5a** and **Figure 5b** show the distribution of the scores of three similarity metrics with and without the use of 3D geometry. The results show that the ability of MEDICO to generate molecules with similar structures to the seed molecules is indeed due to the contribution of 3D geometry.

*Docking score optimization.* Finally, for 3D geometry and molecular docking module, we explore the effect of these two modules on the docking score of generated molecules separately. In each experiment, we generate 1,000 molecules, and the molecules are processed in the same way as above. In addition, we calculate the docking scores of molecules in QM9 as a comparison. The distributions of docking scores of generated molecules under different conditions are shown in **Figure 5c**. In all four cases, the docking scores of generated molecules are improved only in the first set of experiments where both 3D geometry and molecular docking modules are used. This indicates that both 3D geometry and molecular docking are necessary to generate molecules with high docking scores. Furthermore, from the first and last subfigures (the second set and the fourth set) in **Figure 5c**, we can find that 3D geometry is more important than molecular docking in molecular docking experiments.

**Performance on plogP and QED optimization**

Besides the generation of novel and valid molecules, generating molecules with specific properties is also an important task in *de novo* drug discovery. Therefore, the ability to generate molecules with similar properties to drug molecules is also an important capability of molecular generation models. From numerous molecular properties, we select two commonly used to evaluate whether a molecule has the potential to become a drug: (1) **QED** (quantitative estimate of drug-likeness) is a metric of quantifying drug-likeness as a value between 0 and 1. It is not calculated from the molecular structure but is a combination of several molecular descriptors, including molecular weight, logP, topological polar surface area, number of hydrogen bond donors and acceptors, the number of aromatic rings, and rotatable bonds, and the presence of unwanted chemical functionalities[49]. (2) **Penalized-logP (pLogp)**, The full name of logP is the oil-water partition coefficient, which is the logarithmic value of the ratio of the partition coefficient of a compound in n-octanol (oil) and water. It reflects the partitioning of the compound in both oil and water phases. logP value is larger when the compound is more lipophilic, and conversely, smaller when it is more hydrophilic, i.e. the better the water solubility. logP value is related to the dissolution, absorption, distribution, and transport of the drug in the body[52]. But the logP metric has a serious problem, it favors simple alkanes with long chains or large rings. So Kusner et al. proposed Penalized-logP[48] to alleviate this problem. Penalized-logP is the result obtained by subtracting the two penalty terms including synthetic accessibility (SA)[50] and ring size from the logP scores.

We use our well-trained MEDICO to encode the 3D geometry of the molecule as a latent vector $z$, and additionally train a multilayer perceptron to map the latent vector to the property value $y$ of the molecule. In the next step of the generation process, we optimize the latent vector by gradient ascent method, which in turn generates molecules with high property scores. The whole generation process is completed by K steps, where K is a freely settable hyperparameter.
We test the performance of MEDICO with two tasks on property optimization tasks. The first is the optimization of properties for a single molecule. This task aims to optimize molecules with low property scores into molecules with high property scores without significant structural changes. The second is the generation of molecules with higher property scores. To accomplish this task, we select the molecules with the top 500 property scores in the test set for property optimization.

In the single-molecule property optimization task, we randomly select 200 molecules from the test set to optimize their properties. Some of the molecules with improved property scores after MEDICO's optimization are shown in **Figure 6a,b**. It can be seen that after the optimization by MEDICO, the molecules have been greatly optimized in terms of property scores while maintaining the framework structure.

In the task of discovering molecules with high property scores, the top-three property scores of molecules found by MEDICO are shown in **Supplementary Table S3** along with the top-three property scores in the test set. The structural formulae of these molecules are shown in **Figure 6c,d**. We also show in **Figure 6e,f** the top-nine molecular structures in property scores generated by MEDICO. Experimental results show that MEDICO can find molecules with higher property scores than those in the test set. In particular, plogP scores of the molecules generated by MEDICO are much higher than plogP scores of the molecules in the test set. The property scores of the molecules generated by MEDICO are also significantly higher compared to GraphNVP. Further, it is worth noting that the high property score molecules found by MEDICO are structurally very similar in **Figure 6e,f**. For example, molecules with high QED score have five-membered ring or six-membered ring, and molecules with high plogP score are long-chain alkanes. This indicates that MEDICO has successfully found the region in the latent space where molecules with high property scores are located.

**Molecular optimization for inhibitors of SARS-COV-2**
Molecular optimization is a critical step in drug discovery. Substitution of local substructures of drug-like molecules can improve desire properties of drug candidates. Since structurally similar molecules are more likely to have similar physical, chemical and biological properties, replacing key substructures of drug-like molecules with structurally similar substructures can maximize the retention of properties of drug-like molecules and make them better with drug activity.

To test the performance of MEDICO in molecular optimization, we select three substructures of the inhibitor of $M^{pro}$ N3[25] for fragment substitution. We take the selected substructure as a seed molecule and use MEDICO to generate a new structurally similar molecule to replace it for molecular optimization. As a comparison, for each substructure, we randomly select one molecule from the molecules generated by JTVAE[16] for replacement. The sites of the three substructures we choose are illustrated in **Figure 7a**. We adopt a precise docking method CDOCKER[53] to calculate docking results as our metric to evaluate the activity of replaced inhibitor. The results of docking with $M^{pro}$ after replacing three substructures with molecules generated by MEDICO and JTVAE respectively are shown in **Figure 7b,c**. The histograms of -CDOCKER energy are in **Figure 7e**. It can be seen that new inhibitors generated by replacing substructure of N3 with molecules using our MEDICO have a significant increase in negative value of CDOCKER energy compared to N3. This indicates that new inhibitors optimized by MEDICO have stronger inhibitory effect on $M^{pro}$. Substitution using randomly generated molecules from JTVAE result in decrease in the inhibitory capacity of new inhibitors, illustrating the importance of using structurally similar molecules in molecular optimization. The optimization of all substructures achieved better results, proving the robustness of our model. In addition, we try to replace all three substructures simultaneously with MEDICO-generated molecules and observe that new inhibitor has more significant enhancement in -CDOCKER energy. The inhibitor generated by replacing substructures using JTVAE could not be docked due to the implausibility of its structure. This observation further shows the criticality of structural similarity in molecular optimization. Further, we calculate physicochemical properties and ADME (Absorption, Distribution, Metabolism and Excretion) properties of all optimized N3 inhibitors using ADMET Lab 2.0[54] and collate the results in **Figure 7d**. In **Figure 7d**, the green color represents this property is in the standard range, red represents it is out of the standard range, and yellow means it is out of the standard range but close to its edge. The explanation of each property, their reasonable range, and specific values are detailed in supplementary materials. We find that new inhibitors optimized with MEDICO maintain original medicinal properties of N3 better than new inhibitors optimized with JTVAE, demonstrating that

MEDICO can optimize docking ability of inhibitor N3 while better maintaining its drug properties.

In addition, to further validate the robustness of our model on molecular optimization, we take two other inhibitors of M$^{pro}$ reported in the literature, 11a[29] and GC376[55], for a deeper study. We identify two common substructures for these three inhibitors of M$^{pro}$ (**Figure 8a**), so we replace the corresponding substructures of 11a and GC376 with the same molecules as when we replace N3 to observe the changes in their docking energy and drug properties. The experimental results are shown in **Figure 7d** and **Figure 8b-d**. The results demonstrate that MEDICO can still optimize their substructures more effectively on other inhibitors of M$^{pro}$ for docking capacity enhancement and drug property retention, which further proves that our model is highly robust and can be applied to different M$^{pro}$ inhibitors.

## Conclusion

In this paper, we present MEDICO, a **M**ulti-vi**E**w **D**eep generative model for molecule generation, structural optimization, and **I**nhibitor dis**CO**very. Compared to other generative models, MEDICO uses both 2D topology and 3D geometry of molecules and can generate molecules with similar structures to a given molecule. In addition, we give MEDICO chemical priori information through molecular docking, so it could generate molecules that can inhibit SARS-CoV-2. We test the performance of MEDICO on multiple generation tasks, and the experimental results demonstrate that MEDICO can generate molecules that are structurally similar to the seed molecule and (1) efficient and novel (2) have a high docking score with M$^{pro}$ (3) with a high score for the desired property. Molecular optimization for inhibitors of M$^{pro}$ demonstrates that our model can generate new inhibitors with stronger docking capabilities through substructure replacement. For future work, we expect to directly generate accurate 3D conformations of molecules using 3D geometry of molecules and to be able to optimize the given property of molecules.

## Availability of data and materials

All code used in data analysis and preparation of the manuscript, alongside a description of necessary steps for reproducing results, can be found in a GitHub repository accompanying this manuscript: https://github.com/Pang-chao/MEDICO.

## Competing interests

The authors declare that they have no competing interests.

## Author contributions

C.P., L.W. conceived the basic idea and designed the research study. C.P. developed the method. C.P., Y.W., Y.J, drew the figure. RH. W. organized supplementary materials. C.P., R.S. and L.W.

wrote the manuscript.

# Funding

The work was supported by the Natural Science Foundation of China (Nos. 62072329, and 62071278).

# References

1       Chan, H. C. S., Shan, H., Dahoun, T., Vogel, H. & Yuan, S. Advancing Drug Discovery via Artificial Intelligence. *Trends Pharmacol Sci* **40**, 801, doi:10.1016/j.tips.2019.07.013 (2019).

2       Dickson, M. & Gagnon, J. P. The cost of new drug discovery and development. *Discov Med* **4**, 172-179 (2004).

3       Mullard, A. The drug-maker's guide to the galaxy. *Nature* **549**, 445-447, doi:10.1038/549445a (2017).

4       Broach, J. R. & Thorner, J. High-throughput screening for drug discovery. *Nature* **384**, 14-16, doi:10.1038/384014a0 (1996).

5       Yang, S. Y. Pharmacophore modeling and applications in drug discovery: challenges and recent advances. *Drug Discov Today* **15**, 444-450, doi:10.1016/j.drudis.2010.03.013 (2010).

6       McInnes, C. Virtual screening strategies in drug discovery. *Curr Opin Chem Biol* **11**, 494-502, doi:10.1016/j.cbpa.2007.08.033 (2007).

7       Redmon, J., Divvala, S., Girshick, R. & Farhadi, A. in *Proceedings of the IEEE conference on computer vision and pattern recognition.*   779-788.

8       Brock, A., Donahue, J. & Simonyan, K. in *International Conference on Learning Representations.*

9       Vaswani, A. *et al.* in *Advances in neural information processing systems.*   5998-6008.

10      Weininger, D. SMILES, a chemical language and information system. 1. Introduction to methodology and encoding rules. *Journal of chemical information computer sciences* **28**, 31-36 (1988).

11      Gómez-Bombarelli, R. *et al.* Automatic chemical design using a data-driven continuous representation of molecules. *ACS central science* **4**, 268-276 (2018).

12      Ertl, P., Lewis, R., Martin, E. & Polyakov, V. In silico generation of novel, drug-like chemical matter using the LSTM neural network. *arXiv preprint arXiv:.07449* (2017).

13      Grisoni, F., Moret, M., Lingwood, R., Schneider, G. & modeling. Bidirectional molecule generation with recurrent neural networks. *Journal of chemical information* **60**, 1175-1183 (2020).

14      Dai, H., Tian, Y., Dai, B., Skiena, S. & Song, L. in *International Conference on Learning Representations.*

15      David, L., Thakkar, A., Mercado, R. & Engkvist, O. Molecular representations in AI-driven drug discovery: a review and practical guide. *Journal of Cheminformatics* **12**, 1-22 (2020).

16      Jin, W., Barzilay, R. & Jaakkola, T. in *International Conference on Machine Learning.*   2323-2332 (PMLR).

17      Madhawa, K., Ishiguro, K., Nakago, K. & Abe, M. Graphnvp: An invertible flow model for generating molecular graphs. *arXiv preprint arXiv:.11600* (2019).

18      Axen, S. D. *et al.* A simple representation of three-dimensional molecular structure. *Journal of medicinal chemistry* **60**, 7393-7409 (2017).

19      Skalic, M., Jiménez, J., Sabbadin, D., De Fabritiis, G. & modeling. Shape-based generative modeling for

| | de novo drug design. *Journal of chemical information and modeling* **59**, 1205-1214 (2019). |
|---|---|
| 20 | Imrie, F., Bradley, A. R., van der Schaar, M. & Deane, C. M. Deep generative models for 3D linker design. *Journal of chemical information and modeling* **60**, 1983-1995 (2020). |
| 21 | Li, C. *et al.* Geometry-Based Molecular Generation With Deep Constrained Variational Autoencoder. *IEEE Transactions on Neural Networks and Learning Systems* (2022). |
| 22 | Liu, S. *et al.* Pre-training Molecular Graph Representation with 3D Geometry. *arXiv preprint arXiv:.07728* (2021). |
| 23 | Nantasenamat, C., Isarankura-Na-Ayudhya, C., Naenna, T. & Prachayasittikul, V. A practical overview of quantitative structure-activity relationship. *EXCLI Journal* (2009). |
| 24 | Zhang, J. & Chen, H. De Novo Molecule Design Using Molecular Generative Models Constrained by Ligand–Protein Interactions. *Journal of Chemical Information and Modeling* (2022). |
| 25 | Jin, Z. *et al.* Structure of Mpro from SARS-CoV-2 and discovery of its inhibitors. *Nature* **582**, 289-293 (2020). |
| 26 | Morris, G. M. & Lim-Wilby, M. in *Molecular Modeling of Proteins* (ed Andreas Kukol) 365-382 (Humana Press, 2008). |
| 27 | Meng, X.-Y., Zhang, H.-X., Mezei, M. & Cui, M. Molecular docking: a powerful approach for structure-based drug discovery. *Current computer-aided drug design* **7**, 146-157 (2011). |
| 28 | Eberhardt, J., Santos-Martins, D., Tillack, A. F., Forli, S. & Modeling. AutoDock Vina 1.2. 0: New docking methods, expanded force field, and python bindings. *Journal of Chemical Information* **61**, 3891-3898 (2021). |
| 29 | Dai, W. *et al.* Structure-based design of antiviral drug candidates targeting the SARS-CoV-2 main protease. *Science* **368**, 1331-1335 (2020). |
| 30 | O'Boyle, N. M. *et al.* Open Babel: An open chemical toolbox. *Journal of cheminformatics* **3**, 1-14 (2011). |
| 31 | Dolinsky, T. J. *et al.* PDB2PQR: expanding and upgrading automated preparation of biomolecular structures for molecular simulations. *Nucleic acids research* **35**, W522-W525 (2007). |
| 32 | Zang, C. & Wang, F. in *Proceedings of the 26th ACM SIGKDD International Conference on Knowledge Discovery & Data Mining.* 617-626. |
| 33 | Liu, Y. *et al.* Spherical message passing for 3d graph networks. *International Conference on Learning Representations* (2021). |
| 34 | Dinh, L., Krueger, D. & Bengio, Y. in *International Conference on Learning Representations.* |
| 35 | Dinh, L., Sohl-Dickstein, J. & Bengio, S. Density estimation using real nvp. *arXiv preprint arXiv:.08803* (2016). |
| 36 | Hu, W. *et al.* Forcenet: A graph neural network for large-scale quantum calculations. *arXiv preprint arXiv:.01436* (2021). |
| 37 | Klicpera, J., Groß, J. & Günnemann, S. in *International Conference on Learning Representations.* |
| 38 | Ramakrishnan, R., Dral, P. O., Rupp, M. & Von Lilienfeld, O. A. Quantum chemistry structures and properties of 134 kilo molecules. *Scientific data* **1**, 1-7 (2014). |
| 39 | Ruddigkeit, L., Van Deursen, R., Blum, L. C., Reymond, J.-L. & modeling. Enumeration of 166 billion organic small molecules in the chemical universe database GDB-17. *Journal of chemical information* **52**, 2864-2875 (2012). |
| 40 | Liu, Q., Allamanis, M., Brockschmidt, M. & Gaunt, A. Constrained graph variational autoencoders for molecule design. *Advances in neural information processing systems* **31** (2018). |
| 41 | You, J., Liu, B., Ying, Z., Pande, V. & Leskovec, J. Graph convolutional policy network for goal-directed |

molecular graph generation. *Advances in neural information processing systems* **31** (2018).

42  Shi, C. *et al.* in *International Conference on Learning Representations.*

43  Maggiora, G., Vogt, M., Stumpfe, D. & Bajorath, J. Molecular similarity in medicinal chemistry: miniperspective. *Journal of medicinal chemistry* **57**, 3186-3204 (2014).

44  Durant, J. L., Leland, B. A., Henry, D. R. & Nourse, J. G. Reoptimization of MDL keys for use in drug discovery. *Journal of chemical information and computer sciences* **42**, 1273-1280 (2002).

45  Landrum, G. *RDKit: Open-source cheminformatics*, <http://www.rdkit.org/> (2006).

46  Morgan, H. L. The generation of a unique machine description for chemical structures-a technique developed at chemical abstracts service. *Journal of chemical documentation* **5**, 107-113 (1965).

47  Van der Maaten, L. & Hinton, G. Visualizing data using t-SNE. *Journal of machine learning research* **9** (2008).

48  Kusner, M. J., Paige, B. & Hernández-Lobato, J. M. in *International Conference on Machine Learning.* 1945-1954 (PMLR).

49  Bickerton, G. R., Paolini, G. V., Besnard, J., Muresan, S. & Hopkins, A. L. Quantifying the chemical beauty of drugs. *Nature chemistry* **4**, 90-98 (2012).

50  Ertl, P. & Schuffenhauer, A. Estimation of synthetic accessibility score of drug-like molecules based on molecular complexity and fragment contributions. *Journal of cheminformatics* **1**, 1-11 (2009).

51  Pawar, S. S. & Rohane, S. H. Review on discovery studio: An important tool for molecular docking. *Asian J. Res. Chem* **14**, 86-88 (2021).

52  Wildman, S. A., Crippen, G. M. & sciences, c. Prediction of physicochemical parameters by atomic contributions. *Journal of chemical information* **39**, 868-873 (1999).

53  Wu, G., Robertson, D. H., Brooks III, C. L. & Vieth, M. Detailed analysis of grid‐based molecular docking: A case study of CDOCKER — A CHARMm‐based MD docking algorithm. *Journal of computational chemistry* **24**, 1549-1562 (2003).

54  Xiong, G. *et al.* ADMETlab 2.0: an integrated online platform for accurate and comprehensive predictions of ADMET properties. *Nucleic Acids Research* **49**, W5-W14, doi:10.1093/nar/gkab255 (2021).

55  Luan, X.-d. *et al.* Structure basis for inhibition of SARS-CoV-2 by the feline drug GC376. *Acta Pharmacologica Sinica*, 1-3 (2022).

# Figures

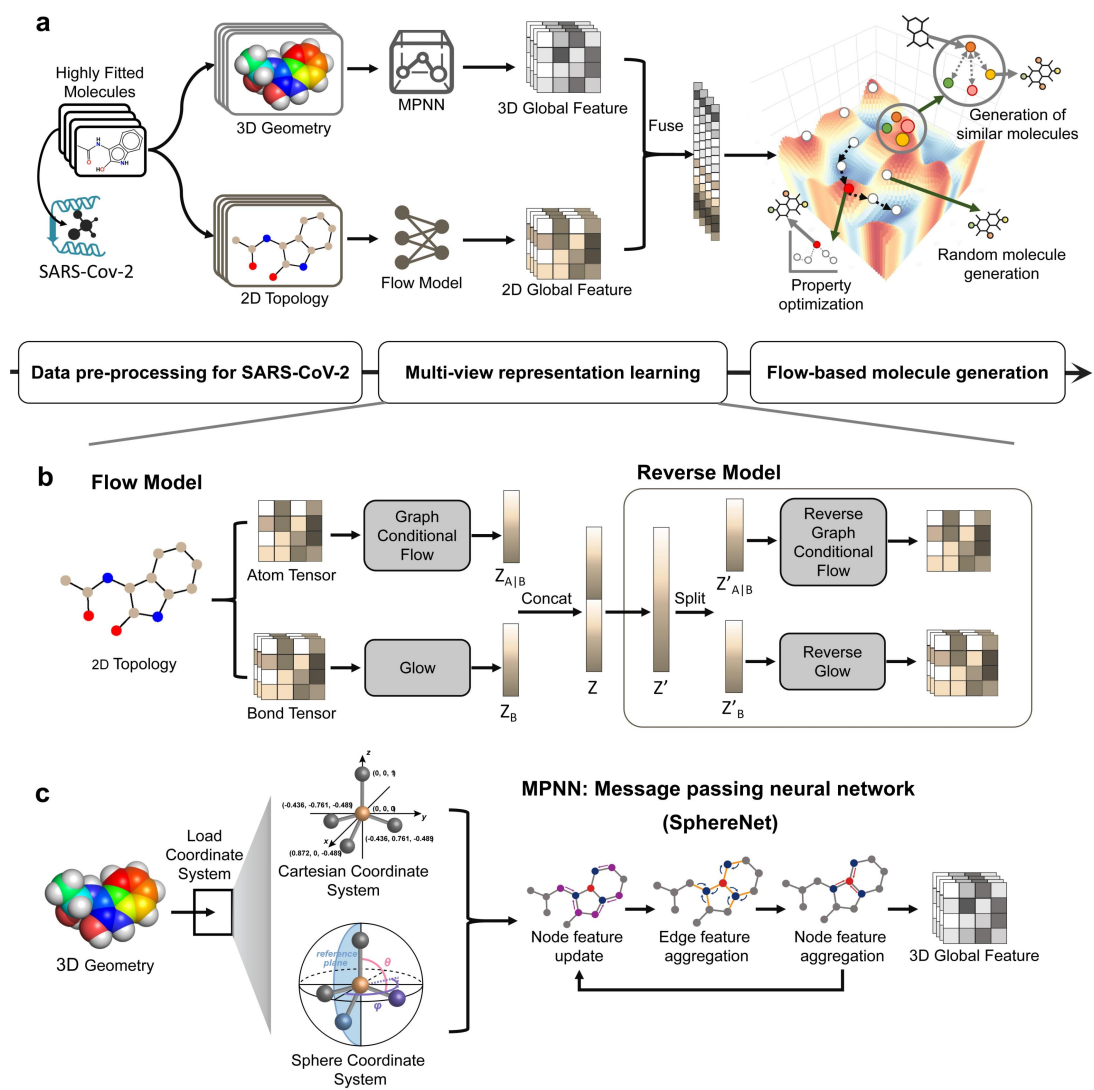

**Fig. 1 | The overview of the proposed MEDICO. a. The framework of MEDICO.** There are three modules: (1) Data pre-processing for SARS-CoV-2, (2) Multi-view representation learning, and (3) Flow-based molecule generation. First, we perform molecular docking of all molecules with M$^{pro}$, and use docking results to assign a weight to each molecule to identify its importance. After molecular docking, the molecules are pre-processed as 2D topology and 3D geometry. Afterward, the resulting 2D topology and 3D geometry are fed into Flow and MPNN, learning 2D and 3D molecular embeddings, respectively. Subsequently, the two embeddings are merged as a joint representation. Finally, in Flow-based molecule generation module, the resulting joint representation is trained with reverse flow model to generate new molecules structurally similar to target molecules. **b.** Workflow of the flow model part in MEDICO. The input molecular graph is first divided into an atomic matrix and a bond tensor, and then these two parts are encoded as latent vectors by Graph conditional flow and Glow respectively. The two latent vectors are combined and fused with 3D features to obtain the joint representation, which is served as input data for Reverse Flow to generate molecules. **c.** The coordinate system conversion and message processing of SphereNet.

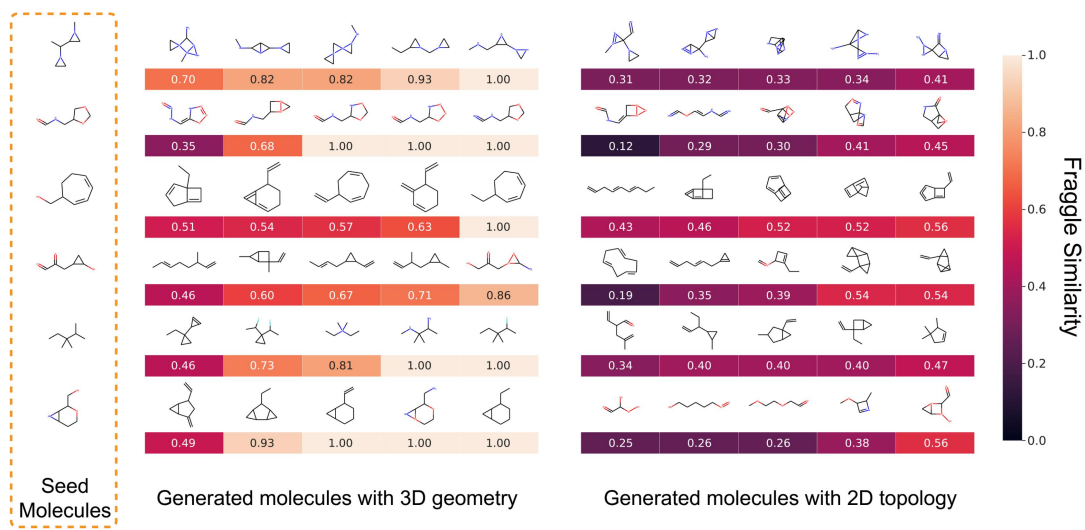

**Fig 2 | Some examples of molecules generated by MEDICO using 3D geometry or 2D topology of seed molecules**. The number below the generated molecule is Fraggle similarity between the generated molecule and the seed molecule. The higher value of Fraggle similarity, the lighter the background color of the number, and the more similar the skeleton structure between the generated molecule and the seed molecule.

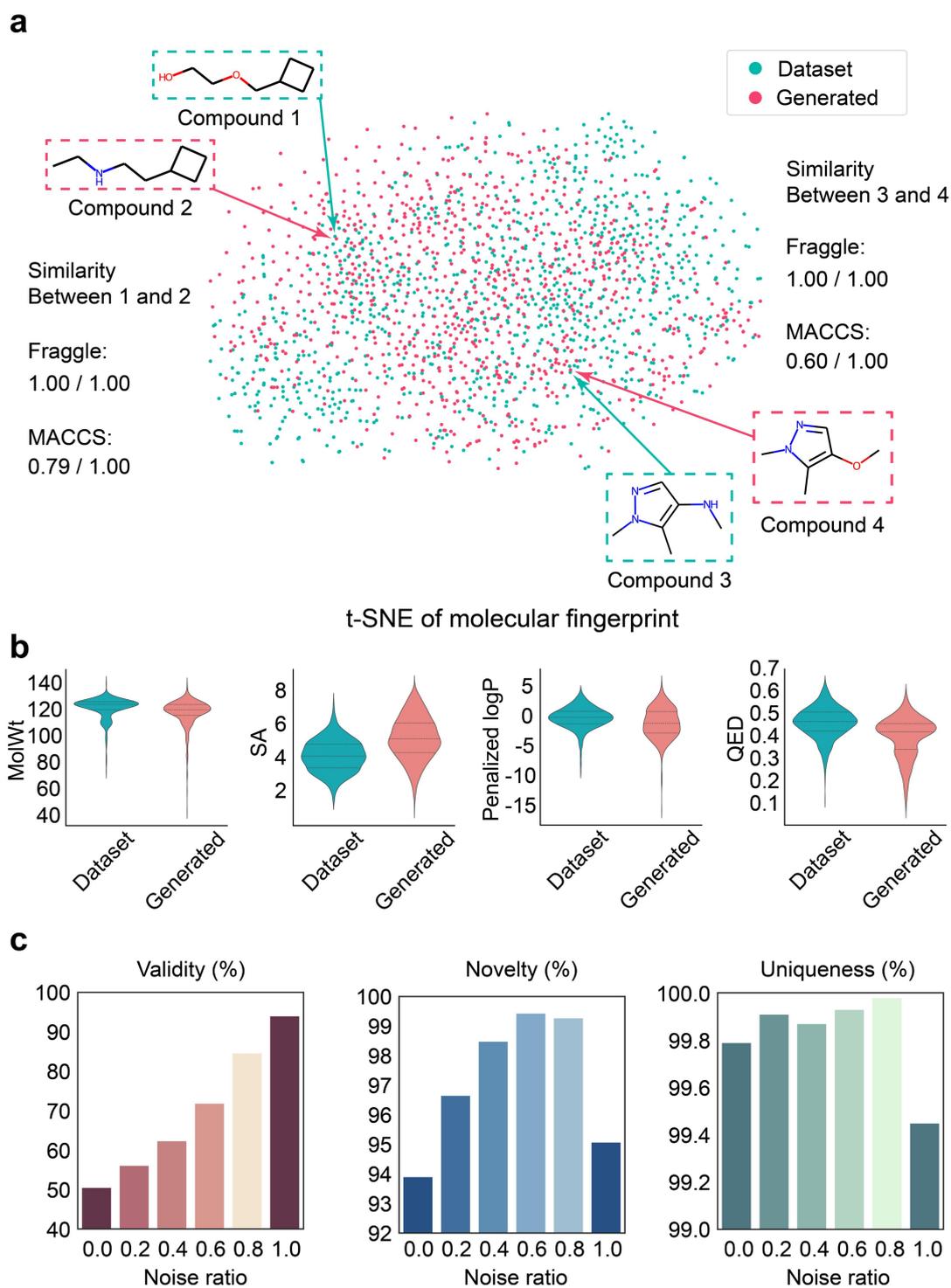

**Fig 3 | Similarity between our generated molecules and molecules in the dataset. a.** t-SNE plots of molecules in QM9 dataset and MEDICO-generated molecules. Green dots represent molecules in the dataset, red dots represent our generated molecules. **b.** The distribution of four property scores of molecules in the dataset and our generated molecules. **c.** The validity, novelty and uniqueness of the generated molecules with different proportions of random noise.

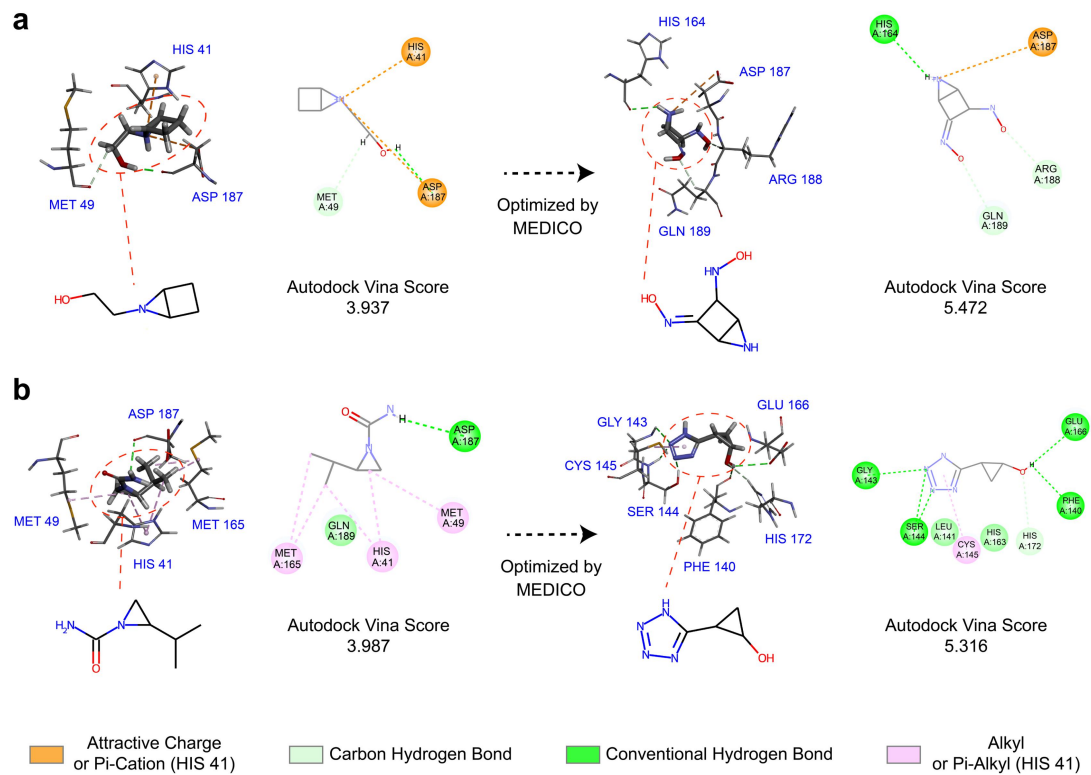

**Fig 4 | Molecular docking results of SARS-CoV-2. a-b.** Two examples of docking score improvement after optimization. In each figure, a 3D schematic of molecular docking is shown on the left, and a 2D schematic is shown on the right.

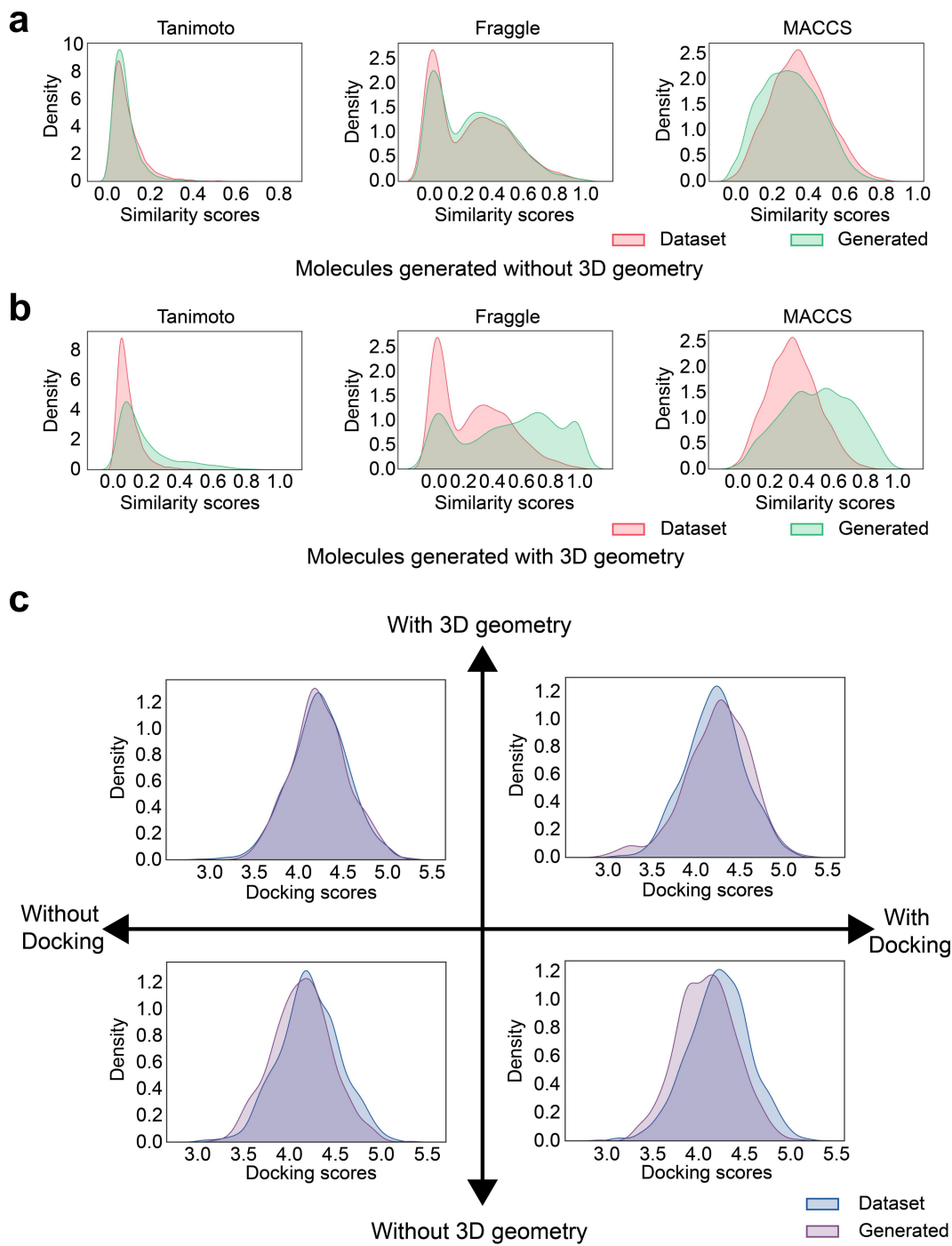

**Fig 5 | Effect of 3D geometry and molecular docking on model performance. a-b.** Distribution of similarity scores of molecules generated without and with 3D geometry. **c.** Comparison of the distribution of the docking scores of the generated molecules with M$^{pro}$ and the docking scores of the molecules in the data set with M$^{pro}$ under four different conditions.

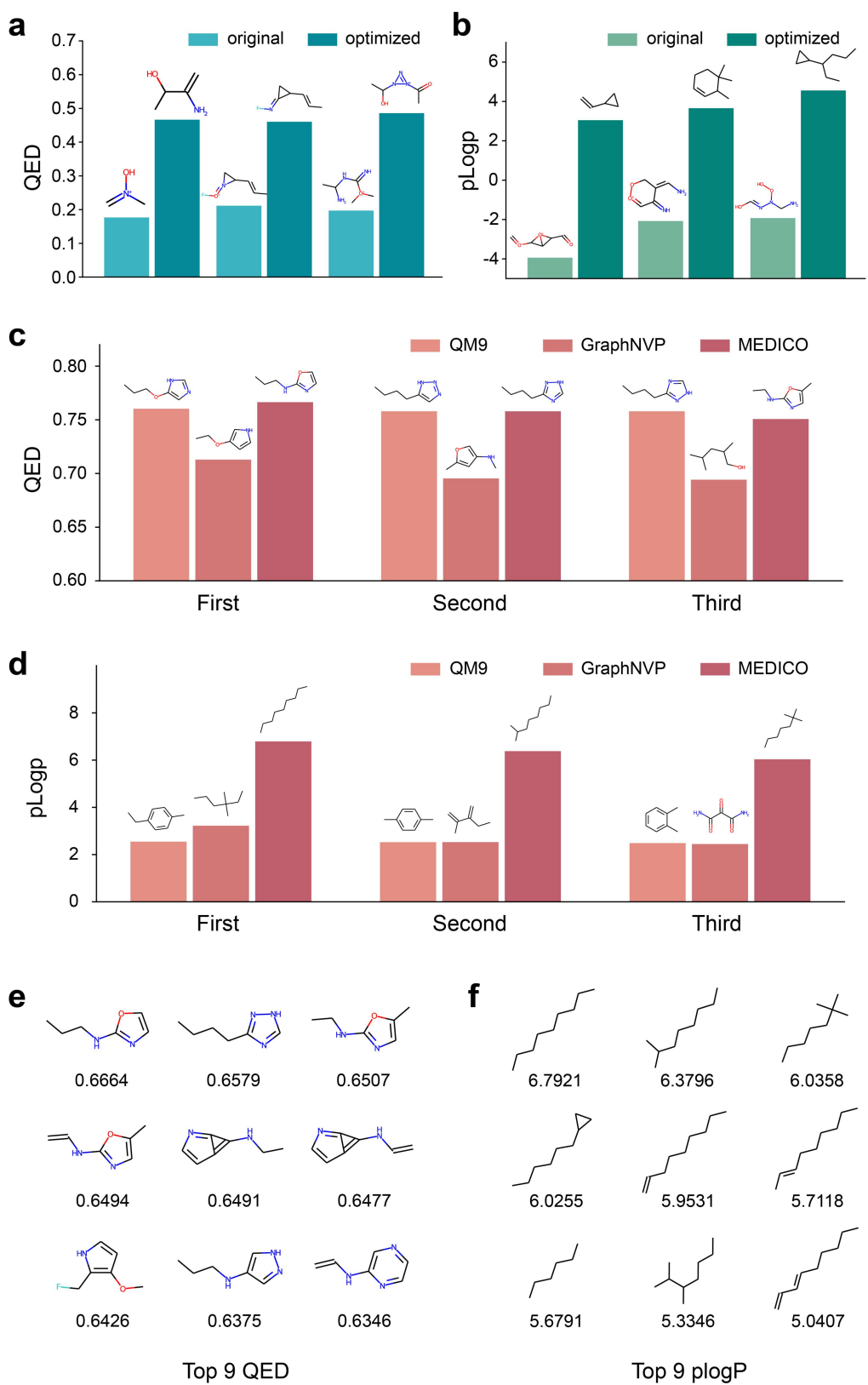

**Fig 6 | Property optimization for pLogp and QED. a-b.** Optimization of properties for a single molecule. we show three examples for each property. **c-d.** Comparison between top-three molecules in property scores generated by GraphNVP and MEDICO and top-three molecules in QM9. **e-f.** The top-nine molecules with the best property scores are found by MEDICO.

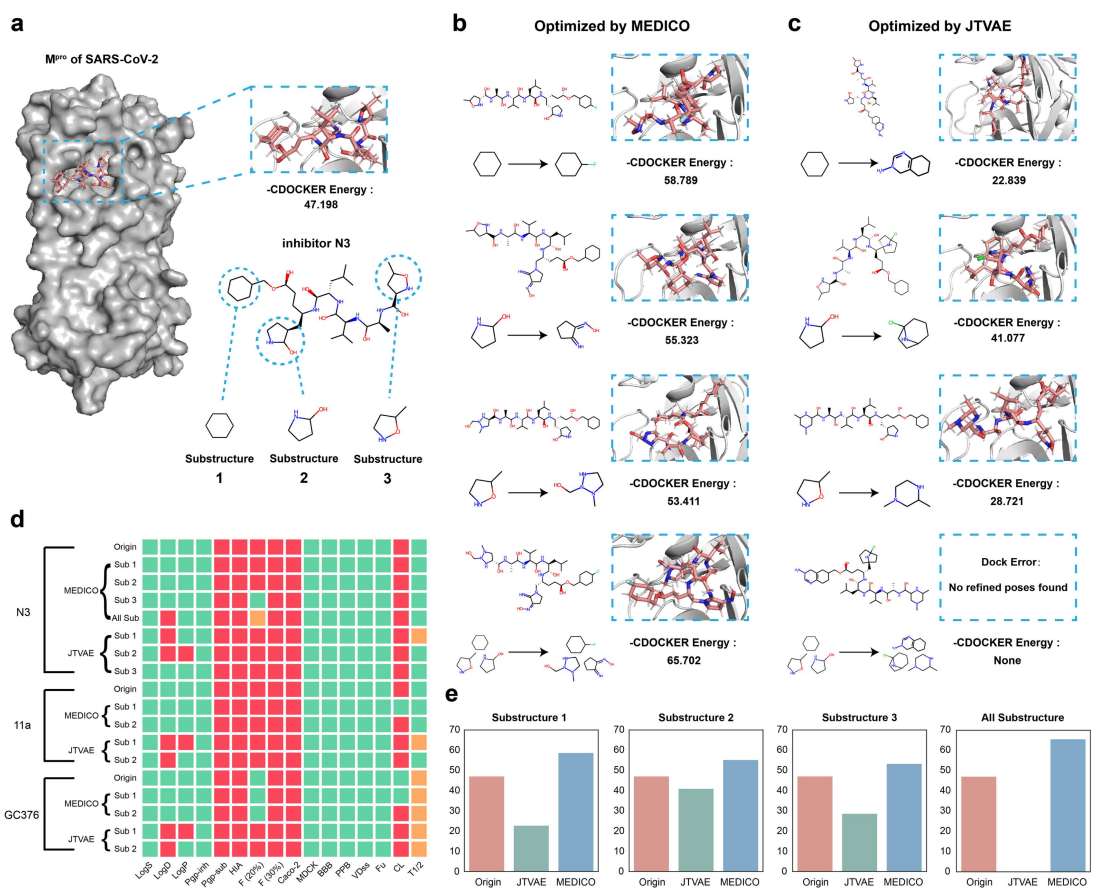

**Fig 7 | Substructure optimization for inhibitor N3. a.** Structure of M[pro], inhibitor N3, and their docking pose. The three selected substructures are circled by dashed lines. **b-c.** Structure of inhibitors optimized by MEDICO and JTVAE and their docking poses and energy scores with M[pro]. **d.** The physicochemical properties and ADME properties of optimized inhibitors. Green color means property value is within reasonable range, red means out of range, and yellow is medium. **e.** Comparison of -CDOCKER energy of original inhibitor and new inhibitors with substructure replacement by MEDICO and JTVAE.

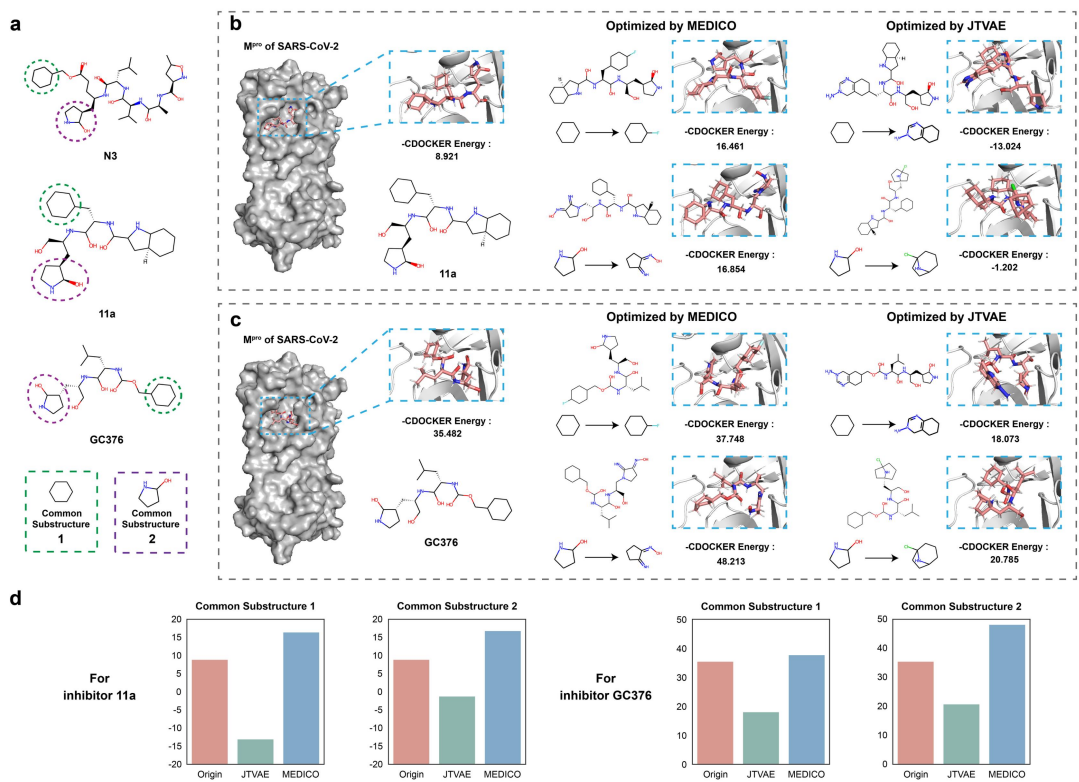

**Fig 8 | Common substructure optimization for inhibitors 11a and GC376. a.** Structure of three inhibitors and their common substructures. **b-c.** Structure of 11a and GC376 inhibitors optimized by MEDICO and JTVAE and their docking poses and energy scores with M$^{pro}$. **d.** Comparison of docking energy of new inhibitors obtained by optimization using MEDICO and JTVAE on two common substructures.

# Tables

Table 1 | Random generation performance on QM9 dataset.

|  | %Validity | %Validity w/o check | %Novelty | %Uniqueness |
|---|---|---|---|---|
| CGVAE | **100** | n/a | 94.35 | 98.57 |
| JT-VAE | 99.86 | n/a | **100** | 96.32 |
| GCPN | **100** | 18.23 | **100** | 87.13 |
| GraphNVP | 50.86 | 50.86 | 88.46 | 97.52 |
| GraphAF | **100** | 46.30 | 91.54 | 99.15 |
| MEDICO | **100** | **94.06** | 95.08 | 99.45 |

Table 2 | Effect of adding and not adding 3D geometry on similarity

|  | **Tanimoto** | **Fraggle** | **MACCS** |
|---|---|---|---|
| QM9 | 10.245% | 27.815% | 35.807% |
| MEDICO (2D) | 10.395% | 30.787% | 34.493% |
| MEDICO (3D) | **19.799%** | **51.966%** | **51.140%** |